\pdfoutput=1

\documentclass[11pt]{article}

\usepackage{xcolor}	
\usepackage{emnlp2021}

\usepackage{times}
\usepackage{latexsym}
\usepackage{booktabs}
\usepackage[T1]{fontenc}

\usepackage[utf8]{inputenc}

\usepackage{microtype}
\usepackage{graphicx}
\usepackage{amsmath,amsfonts}
\usepackage{subcaption}
\newcommand{\norm}[1]{\left\lVert#1\right\rVert}

\newtheorem{proposition}{Proposition}

\title{Sequential Randomized Smoothing for Adversarially Robust Speech Recognition}

\author{Raphael Olivier \and Bhiksha Raj \\
  Carnegie Mellon University \\
  Language Technologies Institute \\
  5000 Forbes Avenue, Pittsburgh PA 15213, USA \\
  \texttt{rolivier@cs.cmu.edu}, \texttt{bhiksha@cs.cmu.edu}}

\begin{document}
\maketitle
\begin{abstract}
While Automatic Speech Recognition has been shown to be vulnerable to adversarial attacks, defenses against these attacks are still lagging. Existing, naive defenses can be partially broken with an adaptive attack. In classification tasks, the Randomized Smoothing paradigm has been shown to be effective at defending models. However, it is difficult to apply this paradigm to ASR tasks, due to their complexity and the sequential nature of their outputs. Our paper overcomes some of these challenges by leveraging speech-specific tools like enhancement and ROVER voting to design an ASR model that is robust to perturbations. We apply adaptive versions of state-of-the-art attacks, such as the Imperceptible ASR attack, to our model, and show that our strongest defense is robust to all attacks that use inaudible noise, and can only be broken with very high distortion.
\end{abstract}

\section{Introduction\footnote{For reproducibility purposes our code and models are available at https://github.com/RaphaelOlivier/smoothingASR}}
\label{sec : intro}
\subsection{The threat of adversarial attacks on ASR}
In recent years, Automatic Speech Recognition (ASR) has been transitioning from a topic of academic research to a mature technology implemented in everyday devices. AI voice assistants are becoming increasingly popular, and ASR models are being implemented in cars, smart TVs, and various devices within the \textit{Internet of Things}. Therefore, challenges to the security of these models, as explored in several recent articles, are also transitioning from academic curiosities to real-world threats. 

One of these major security threats is vulnerability to \textit{adversarial attacks} \cite{Szegedy14}: perturbations of inputs to any model that, while nearly imperceptible to human senses, have considerable effects on its outputs. Such attacks can enable a malicious party to discretely manipulate models' behaviors and cause them to malfunction while escaping human observation. For instance, when applied to voice assistants, adversarial attacks can potentially lead to privacy breaches by successfully soliciting arbitrary sensitive information. They could also fool an ASR system to believe an audio input contains hateful content, and have it automatically rejected from platforms or its author banned.

For several years now, adversarial attacks have been an active research field that crosses nearly every application of Machine Learning. One of the main objectives of the field is to defend AI models against such attacks without impacting their performance (on regular data) heavily. 
\subsection{Limitations of current defenses}
This research has taken the form of an arms race, where the attacker has the upper hand: whenever a defense was proposed \cite{samangouei2018defensegan}, a stronger or \textit{adaptive} attack was developed to counter it \cite{athalye18}.
Some recent works seem to have partially broken this trend by proposing defenses with precise claims, that are \textit{optimal} in a specific sense or \textit{certified} against specific classes of attacks. These defenses mostly fit into three categories: \textit{Adversarial training} using strong attacks like PGD \cite{madry18}, \textit{Convex relaxations} of the adversarial training objective \cite{wong2018provable}, and noise-based \textit{randomized smoothing} methods \cite{cohen19}.

These defenses, however, have all been proposed on classification tasks, and their extension to ASR is not trivial. Adversarial training, which is already time and resource-consuming for classification, is even harder to use in speech recognition where strong attacks are much longer to compute. Convex relaxation is heavily architecture-dependent, and the use of recurrent networks or different activations makes it hard to adapt to ASR. Randomized smoothing is more promising because its simple Gaussian noise-addition method makes it, in principle, usable on any model without concerns for how the attacks are computed. In reality however, there are still major challenges. ASR models are typically more susceptible than classification models to the same amount of Gaussian noise. Besides, to retain good model performance with large amounts of random noise, smoothing methods require running multiple iterations of the randomized model and using a majority vote on the outputs. When evaluating sentences in the English alphabet, the set of outputs is exponentially large in the length of the output, and majority vote is unlikely to estimate accurately the most probable one within a reasonable number of iterations.

\subsection{Our contributions}
Overcoming these challenges is the object of our work. Since general-purpose machine learning defenses have significant limitations when applied to speech, we improve them by leveraging the tools developed by the Speech Processing community. To use randomized smoothing on ASR while retaining good clean performance, we consider speech enhancement methods to make the defended model more accurate on Gaussian-augmented inputs. We also replace the majority vote with a strategy based on the ``Recognizer Output Voting Error Reduction'' (ROVER) \cite{rover} method. Depending on whether we apply training data augmentation, we provide both an off-the-shelf defense and one that requires specific fine-tuning.

We apply our defenses to a DeepSpeech2 and a Transformer model, trained and evaluated on the LibriSpeech dataset. We test them against strong attacks like the CW attack \cite{Carlini18}, the Imperceptible ASR Attack \cite{qin19}, and the (untargeted) PGD attack \cite{madry18}. We run adaptive versions of these attacks to avoid obfuscation effects. Our best model shows strong robustness against these attacks: to achieve partial transcription of the target sentence, attack algorithms require $10$ times larger perturbations. Under equal noise distortions, the Word-Error-Rate (WER) on the ground truth under denial-of-service attacks improves by 30 to 50\% for our model compared to the baseline.

\section{Related Work}
\label{sec : relwork}
\subsection{Attacks}
Numerous general adversarial attacks have been proposed in the past \cite{Szegedy14,goodfellow15,Moosavi16,Carlini17,madry18}. A few others specifically targeted audio inputs: the earliest was the ultrasonic-based DolphinAttack \cite{DolphinAttack} and the Houdini loss for structured models \cite{houdini}, followed by the effective and popular Carlini\&Wagner (CW) attack for audio \cite{Carlini18}. Other works have extended the state-of-the-art with over-the-air attacks \cite{CommanderSong,Yakura19,li19} and black-box attacks that do not require gradient access and transfer well \cite{Abdullah21}. A recent line of work has improved the imperceptibility of adversarial noise by using psychoacoustic models to constrain the noise rather than standard $L_2$ or $L_\infty$ bounds \cite{Szurley2019PerceptualBA,Schoenherr2019,qin19}.

\subsection{Defenses}
While a large number of defenses against adversarial attacks have been proposed (\citet{Papernot15,Buckman18,samangouei2018defensegan} are just examples), the vast majority have been broken using either a strong or an adaptive attack \cite{Carlinidetection,athalye18}. Only a handful of defense families have stood the test of time. One is adversarial training, in the form proposed by \citet{madry18} as well as more recent variations \cite{Wong2020FastIB,Tramer19}. Noise-based, smoothing methods are another \cite{cao17,Li18,Liu18,Lecuyer19,cohen19}. Finally, some methods prove robustness by investigating a relaxation of the adversarial objective \cite{Gowal,wong2018provable,Mirman} or an exact solution \cite{KBD+17,Bunel}.

Efforts to adapt adversarial training or relaxation methods to ASR have been limited so far: \citet{Sun18} have used training for speech based on the FGSM attack, which is simpler but not nearly as robust as PGD training. Most proposed ASR defenses such as MP3 compression \cite{Das} or quantization \cite{Zhuolin19} have shown the same weakness as above to adaptive attacks \cite{Subramanian19}. Exploiting temporal dependencies in speech to detect adversarial manipulations \cite{Zhuolin19} is a promising line of work. However, at best it only enables the user to \textit{detect} these modified inputs. \textit{Reconstructing} the correct transcription is an entirely different challenge, and our objective in this work. 

\subsection{Randomized smoothing for ASR}
Some noise-based defenses for \textit{audio classification} have been proposed: \citet{Subramanian19} for instance use simple white noise as a defense mechanism. This is a straightforward extension of randomized smoothing to another classification setting. 

Regarding specifically ASR, the only existing randomized smoothing works we are aware of are \citet{mendes}, who propose an adaptation of the noise distribution to psychoacoustic attacks, and the recent \citet{zelasko21}. The latter in particular thoroughly explores the effects of Gaussian smoothing on DeepSpeech2 and the Espresso Transformer. However, their work on making these models more robust to white noise is limited to Gaussian augmentation in training. Specifically, they do not explore the issue of voting on transcription and resort to one-sentence estimation (see Section \ref{sec : votenaive}), which limits the amount of noise they can use, and therefore the radius of their defense. Besides, they do not use adaptive attacks (Section \ref{sec : adapt}) which makes their evaluation incomplete. To our knowledge, we propose the first \textit{complete} (randomization, training and vote, evaluated on adaptive attacks) version of randomized smoothing for Speech Recognition.

\section{Adversarial attacks on Speech Recognition}
\label{sec : attack}
As in previous work, we evaluate our defenses against white-box attacks, that can access model weights and their gradients and are aware of the defenses applied. Provided with an input, these attacks will run gradient-based iterations to craft an additive noise. They are the hardest attacks to defend against, and a great metric to evaluate defenses that will carry over well to more practical attacks run over-the-air, without gradient access or in real-time \cite{li19, Abdullah21}.

We consider two threat models. \textit{Untargeted attacks} generate a small, additive adversarial noise that causes a denial-of-service (DOS) by altering drastically the transcription. \textit{Targeted attacks} on the other hand craft an additive noise that forces the model to recognize a specific target of the attacker's choice, such as "OK Google, browse to evil.com" \cite{Carlini18}. As their objective is more precise than simple denial-of-service, targeted attacks typically require slightly larger perturbations.

We now present the specific attacks that we use. Perturbed samples for all these attacks are provided as supplementary material.

\subsection{Untargeted attacks}
\paragraph{Projected Gradient Descent} The PGD attack \cite{madry18} crafts a noise $\delta$ that generates mistranscriptions by maximizing the loss under its perturbation budget. It optimizes the objective $$max_{|\delta|_\infty\leq\epsilon}L(f(x+\delta),y)$$ using Projected Gradient Descent\footnote{or Ascent, in this case}: it takes gradient steps that maximize the loss $$\delta_n \leftarrow \delta_{n-1} + \eta L(f(x+\delta_{n-1}),y)$$ and projects $\delta_n$ on the ball of radius $\epsilon$ after each iteration. We use 50 gradient steps when running this attack. 

Rather than fixing a value for $\epsilon$ over all sentences, it is more interesting to bound the relative amount of noise compared to the input, that is the \textbf{signal-noise-ratio} (SNR)  expressed in decibels:
$$SNR(\delta,x) = 20*log_{10}(\frac{\norm{x}_2}{\norm{\delta}_2})$$ When running PGD attacks, we set an SNR threshold, then derive for each utterance the $L_\infty$ bound $\epsilon=\frac{\norm{x}_2}{10^{\frac{SNR}{20}}}$.

\subsection{Targeted attacks}

\paragraph{Carlini\&Wagner attack} The CW attack \cite{Carlini18} is a targeted attack, specifically designed against CTC models. For a specific attack target $y_T$ it minimizes the objective: $$L(f(x+\delta),y_T) + \lambda\norm{\delta}_2$$
This attack is \textit{unbounded}, which means it does not fix a threshold for how large $\delta$ should be. Instead, it will regularly update its regularization parameter $\lambda$ to find the \textit{smallest successful} perturbation. Therefore, the most interesting metric to evaluate a model under this attack is the SNR it achieves. \citet{Carlini18} report SNRs between 30 and 40dB on the undefended DeepSpeech2 model.

To run this targeted attack, we fixed 3 target sentences of different lengths, constant in all our experiments. We try to perturb each input utterance until the model generates one of the targets (the one of closest length). For example, all utterances of less than 3-8 words are perturbed to predict the target \textit{"Really short test string"}.

\paragraph{Imperceptible ASR attack}\label{sec : imp} This attack proposed by \citet{qin19} is a variation of the CW attack for ASR \cite{Carlini18} that adds a second objective, where masking thresholds are computed on specific frequencies, to make the noise less perceptible by the human ear. The Imperceptible attack does not improve the SNR budget of the CW attack, only how these examples are perceived by the listener under a fixed budget. Therefore reporting its results is superfluous. We however provide samples generated by this attack along with this article.

\subsection{Adaptive attacks against defended models} \label{sec : adapt} Our defenses use randomized (Gaussian smoothing) and non-differentiable (speech enhancement) preprocessing steps. As \cite{athalye18} have shown such elements can obfuscate the gradients and lead authors to wrongfully assume that a defense is robust. We follow the recommendations of that paper, and adapt our attacks to alleviate these effects, using two techniques: \begin{itemize}
    \item \textit{Straight-through estimator}: when flowing gradients through the non-differentiable preprocessing module, we approximate its derivative as the identity function.
    \item \textit{Expectation over Transformation}: since our model is stochastic, rather than just applying backpropagation once to compute gradients, we average the gradients returned by 16 backpropagation steps.
\end{itemize}

We illustrate the need for such attacks in appendix \ref{apx : adapt}.

\section{Randomized smoothing for speech recognition}
\label{sec : rsasr}

\subsection{Randomized smoothing for classification}
\label{sec : rscl}
The idea of defending models against attacks by adding random noise to the inputs was formalized and generalized in \citet{cohen19} for classification. The idea is to replace the deterministic classifier $f:\mathbb{R}^d \rightarrow \{1,2,...,m\}$ with the smooth classifier: $$g(x)=argmax_{k\in \{1,2,...,m\}}\mathbb{P}(f(x+\epsilon)=k)$$ with $\epsilon \sim \mathcal{N}(0,\sigma^2I)$. More precisely, since classifier $g$ cannot be evaluated exactly, it is estimated with a form of Monte Carlo algorithm: many noisy forward passes are run and majority vote determines the output label.



The underlying reasoning behind this method is that given a small perturbation $\delta$, and a standard deviation $\sigma >> \norm{\delta}_2$, probability distributions for $x+\epsilon$ and $x+\delta+\epsilon$ are very ``close'' by standard divergence metrics. Therefore discrete estimators built around these distribution have equal value with high probability. So if an attacker crafts an adversarial perturbation $\delta$, it will have a very small chance of changing the output of $g$.

When using randomized smoothing, the biggest challenge is to retain good performance on very noisy data. One can see this defense as a way to shift the problem from adversarial robustness to white noise robustness. This is typically done with data augmentation during training.
\subsection{Extension to variable-length data}
\label{sec : rsvarlen}

The variable length of speech inputs is not an issue to use randomized smoothing. The main consequence is that the $L_2$ norm of a perturbation scales with the utterance length. Since Signal-Noise Ratio is normalized by utterance length, this does not affect our experiments.

A bigger problem is the nature of the text transcriptions output by the model. The number of possible outputs is exponential in the length of the input, and the probability mass of each of them under noisy inputs is extremely small. Therefore, majority vote cannot estimate the probabilities of the transcriptions in practice, as we discuss in Section \ref{sec : votenaive}.

However, the reasoning that Gaussian distributions centered on an utterance $x$ and an adversarially perturbed one $x+\epsilon$ are close is still valid. This tends to show that the overall noise-additive method still makes natural and adversarial points similar from the model's perspective. 

\subsection{Gaussian noise-robust speech recognition models}
\label{sec : gausasr}
As mentioned above, when using randomized smoothing it is critical to retain good performance on Gaussian-augmented inputs. With ASR models this is not a trivial objective. We consider the following techniques to achieve this goal.

\paragraph{Augmented Training}
Rather than training a neural ASR model entirely on Gaussian-augmented data, we used a pretrained model on clean data that we fine-tune with Gaussian augmentation for one epoch. We find that it helps training converge and leads to similar or better results on noisy data in a much shorter time.

\paragraph{Speech enhancement} Speech enhancement algorithms help improve audio quality. After adding Gaussian noise, we can use enhancement to restore the original audio quality. We tried multiple standard enhancement methods and found apriori SNR estimation (ASNR) \cite{asnr} to be most effective. Neural methods such as SEGAN \cite{segan} did not reach the same performance (in terms of WER in the end-to-end ASR pipeline), most likely because these models are tailored to real-world-like noise conditions and are not trained on Gaussian noise.  On image tasks, \cite{Salman20DS} have successfully used a trained denoiser alongside randomized smoothing. It is possible that such a model trained on Gaussian noise would be effective on speech as well. It is however not certain at all that it would improve our enhancement results: \citet{segan} argue that they outperform first-order filters like ASNR specifically for complex noise conditions.


\section{Voting strategies on text outputs}
\label{sec : vote}
Even with augmented training and/or enhancement, when feeding noisy inputs to ASR models the output distribution has high variance. Running multiple forward passes and "averaging" the outputs can help reduce that variance and improve accuracy. But this requires a good voting strategy on text outputs.

We first discuss some elementary strategies and their potential drawbacks, then describe the ROVER-based vote that we use. All of these strategies are empirically compared in Section \ref{sec : noisyres}. We denote the sampled transcriptions as $t_1,...,t_n$ and $t$ is our final transcription.
\subsection{Baseline strategies}
\label{sec : votenaive}
\paragraph{One-sentence estimation}
A solution that has the merit of simplicity is to not vote at all. Using only one input, we can hope that the sentence we get is "close" to the most probable sentence (in terms of Word-Error Rate for instance) and just return it as our output. This is the strategy used by \citet{zelasko21}.

\paragraph{Majority vote}
Following the original randomized smoothing defense, we can vote at sentence level: $t = argmax_{t'\in\{t_1,...,t_n\}}card(\{i/t_i=t'\})$

Designed for classification, majority vote is not adapted to probabilistic text outputs. The set $T$ of all possible transcriptions is infinite, and even with our most stable models and a relative noise of -15 dB, 100 noise samples typically output 100 different transcriptions. Even without setting up rigorous statistical tests, it is clear that outputting the likeliest transcription, or just a "likelier than average" one, with high probability would require thousands of ASR iterations, which in practice is not feasible. In other words, majority vote is barely better than one-sentence estimation, for a high computation cost.

\paragraph{Statistics in the logits space}
For a given input utterance length, some ASR architectures, such as CTC-trained models \cite{ctc}, first generate fixed-length logits sequences $l_1,...,l_n$, then apply a best-path decoder $d$ to generate transcriptions $t_i=d(l_i)$. It is then possible to aggregate these logits over the random inputs, for example by averaging them, then to apply the decoder: 
$t = d(\frac{1}{n}\sum_il_i)$.

One potential issue with this strategy as a defense is that it distances itself from the randomized smoothing framework, where the use of \textit{discrete} outputs to vote on is critical. To get a concrete idea of how this could be a problem, one should remember that adversarial examples can be generated with \textit{high confidence} (aka very large logits). Such a phenomenon could disrupt the statistic by over-weighting the fraction of inputs that are most affected by an adversarial perturbation.

\begin{table*}[h]
    \centering
    \begin{tabular}{@{\extracolsep{2pt}}ll|l|l|l@{}}
    \hline
       & Model  & $\sigma=0$ & $\sigma=0.01$  &  $\sigma=0.02$ \\
      \hline
      \textit{Baseline} &   Deepspeech2  & 9.7  &  27 & 63  \\
       &   Transformer  &  5.7 & 12  & 35  \\
      \hline
       &   Deepspeech2$+\operatorname{\sigma-AUG}$  &  - &  11 & 16  \\
      \textit{Enhancement,} &   Deepspeech2$+\operatorname{ASNR}$  & -  &  17 & 33  \\
      \textit{augmentation} &   Deepspeech2$+\operatorname{\sigma-AUG}+\operatorname{ASNR}$  & -  & 12  & 18  \\
        &   Transformer$+\operatorname{ASNR}$  & -  & 9.1  & 19  \\
        \hline
       &   Deepspeech2$+\operatorname{Maj-100}$  & -  & 27  & 69  \\
       &   Deepspeech2$+\operatorname{Avg-100}$  &  - & 25  & 62  \\
      \textit{Voting} &   Deepspeech2$+\operatorname{ROVER-8}$  &  - & 21  & 52  \\
       &   Deepspeech2$+\operatorname{ROVER-16}$  &  - & 20  & 50  \\
       &   Deepspeech2$+\operatorname{ROVER-32}$  &  - & 20  & 50  \\
      \hline
       
      \textit{Proposed models}
       &   Deepspeech2$+\operatorname{ASNR}+\operatorname{ROVER-16}$  & -  &   14 & 26  \\
       & Deepspeech2$+\operatorname{\sigma-AUG}+\operatorname{ROVER-16}$  & -  &  9 & \textbf{12}  \\
       &   Transformer$+\operatorname{ASNR}+\operatorname{ROVER-16}$  & -  &  \textbf{8.1}  & 15  \\
       \hline
    \end{tabular}
    \caption{Word Error Rate (\%) for Deepspeech2 on the LibriSpeech clean test set under various defenses on \textit{clean} utterances when adding Gaussian noise of deviation $\sigma$. $\operatorname{\sigma-AUG}$ stands for Gaussian augmentation of deviation $\sigma$ in training - the same deviation used at inference. $\operatorname{ASNR}$ means A priori SNR filtering of inputs. $\operatorname{Maj-N}$,$\operatorname{Avg-N}$,$\operatorname{ROVER-N}$ refer to majority vote, logits averaging and ROVER voting strategies, using $N$ forward passes.
    \label{tab : clean}}
\end{table*}
\subsection{ROVER}
\label{sec : rover}
The Recognizer Output Voting Error Reduction (ROVER) system was introduced by NIST in 1997, as an ensembling method that mitigates the different errors produced by multiple ASR systems. Contrary to majority vote it works at the word level rather than the sentence level, by selecting at each position the word present in the most sentences. ROVER should be fed the time duration of each word in the audio space, which we can extract using audio-text alignment information provided by the ASR models (see Section \ref{sec : model}). We use ROVER as a black-box script and understanding its inner behavior is not absolutely necessary to follow this work, however we provide a high-level explanation of this algorithm in Appendix \ref{apx : rover}

In our work, we introduce an alternative use of ROVER, as a voting system on the text outputs of the same probabilistic model rather than for ensembling multiple models. We mostly use it as a black box, by feeding to ROVER multiple output sequences.

The main drawbacks of this method lie in the time penalty of the voting module when using a large number of inputs. We further discuss that limitation in Section \ref{sec : anavote}

\section{Experiments}
\label{sec : exps}
\subsection{Dataset}
\label{sec : data}
We run all our experiments on the 960 hours LibriSpeech dataset \cite{librispeech}, and report our results on its test-clean split. As adversarial attacks can take a considerable amount of time to compute, we evaluate attacks on the first 100 utterances of this test set.

\subsection{Models}
\label{sec : model}

We test our smoothing methods on two model architectures:

\begin{itemize}
    \item The CTC-based DeepSpeech2 \cite{deepspeech2}, a standard when evaluating adversarial attacks on ASR since \citet{Carlini18}. We pretrain it on the clean LibriSpeech training set. As discussed above, we fine-tuned it on Gaussian-augmented data for one epoch, using always the same deviation used at inference for smoothing. For decoding we use greedy search, as we find that increasing the beam size has very little impact on WER for this model. This is a relatively lightweight model that we use for ablation experiments. The CTC decoder provides frame alignments for each transcription character: we use them to infer word duration (needed for ROVER) with good precision.
    
    \item A more recent Transformer architecture. We adapt the Espresso implementation \cite{wang2019espresso} to our code. Training and hyperparameter search for transformer models can be computationally expensive, and reaching state-of-the-art word-error-rate is unnecessary in this work. For those reasons we keep all the hyperparameters of Espresso's "Librispeech Transformer" architecture, and do not fine-tune this model on Gaussian noise. We also only run untargeted attacks on the transformer model, as targeted attack algorithms and implementations are usually model specific (and most often proposed against DeepSpeech2). We do not know of any available implementation of a targeted attack on ASR transformer models.
    
    This transformer implementation does not output character alignment. It however provides word-level attention scores with the encoded audio. We can align each word with the highest-scoring audio vector, and obtain a word-level alignment. This method is less precise than with DeepSpeech2 and can likely be improved.
    
\end{itemize}

\subsection{Defenses}
\label{sec : def}
Our models are defended with Gaussian noise, ASNR enhancement, voting strategies or a combination of all of these. The noise deviation is set to $\sigma=0.1$ or $\sigma=0.2$ depending on the experiments, which corresponds for the vast majority of utterances to a signal-noise ratio in the 10-14dB and 7-11dB respectively.

Against adversarial examples, we compare our models with each other, as well as with the undefended DeepSpeech2 model and a baseline defense using MP3 Compression \cite{Carlini18}.

\subsection{Evaluation metrics}
\label{sec : metric}
We evaluate our models under untargeted attacks with \textbf{Word Error Rate} (WER) which is the word-level edit distance between two transcriptions normalized by the length of the target. In case of large mistakes we upper bound WER values to 100\% (while the real value can be greater if the generated sentence is longer than the target). We report this WER on the ground truth: the lower the WER the better the defense. 

With targeted attacks, we also report Word-Error-Rate, both on the ground truth and the attack target. However, since the CW attack is unbounded, if applied with good hyperparameters it always succeeds in forcing our model to deviate from the ground truth (high WER) and predict the attack target (low WER). Therefore these metrics mainly have value as a sanity check to make sure we run the attack correctly. To measure whether a defense is effective against CW, our primary evaluation metric is the \textbf{Signal-Noise-Ratio} (SNR), which quantifies the \textit{amount} of noise the attack generates to achieve its objective. An SNR above $20-25$ would typically be hard to perceive for a human.

\begin{figure*}[h]
\centering
  \centering
  \includegraphics[trim=0 0 2 2, clip,width=0.8\linewidth]{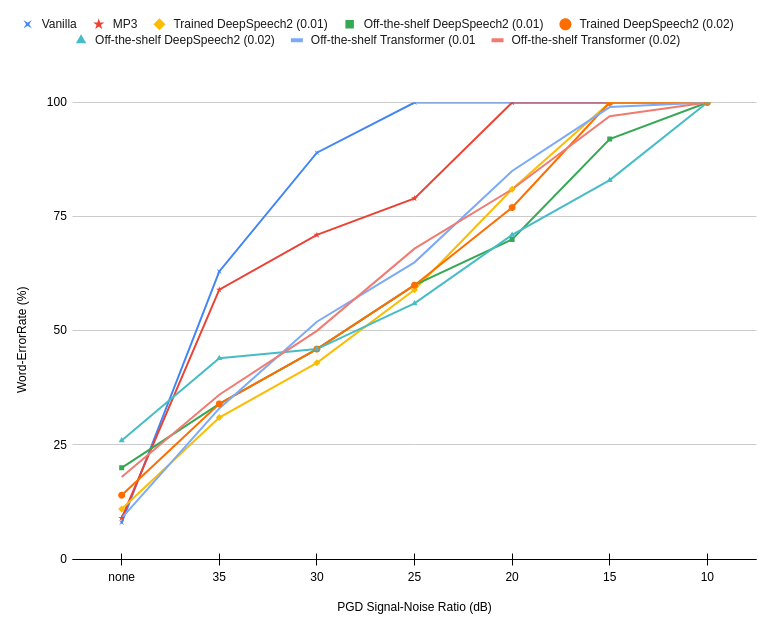}
\caption{WER achieved against PGD attacks by the baselines (Vanilla model, MP3 compression) and the proposed trained (DeepSpeech2 with Smoothing, augmented training and ROVER) and off-the-shelf (DeepSpeech2/Transformer with Smoothing, ASNR and ROVER) defenses, with Gaussian deviations $0.01$ and $0.02$. We plot the results when varying the PGD SNR bound: a lower SNR means a larger perturbation.}
\label{fig : attack}
\end{figure*}
\begin{table}[h]
    \centering
    \resizebox{\linewidth}{!}{
    \begin{tabular}{@{\extracolsep{1pt}}l|l|lll@{}}
    \hline
        Model  & No  & \multicolumn{3}{c}{CW attack}\\
        & attack & GT & TGT & SNR\\
      \hline
      Vanilla & 8 & 100 & 0 & 27dB  \\
        MP3 compression & 9 & 100 & 3 & 16dB \\
      \hline
        Trained ($\sigma=0.01$) & 11 & 100 & 8 & 14dB \\
        Off-the-shelf ($\sigma=0.01$) & 17 & 100 & 4 & 10dB \\
      \hline
        Trained ($\sigma=0.02$) & 14 & 100 & 6 & 8dB \\
        Off-the-shelf ($\sigma=0.02$) & 31 & 100 & 6 & 5dB \\
       \hline
    \end{tabular}}
    \caption{Word Error Rate (\%) for Deepspeech2 on the first 100 utterances of the LibriSpeech clean test set, for clean inputs and under the CW attack for the baselines and the proposed trained and off-the-shelf defenses, with Gaussian deviations $0.01$ and $0.02$.
    We report both the WER on the ground truth (GT) and the attack target (TGT), and the SNR required to achieve it.
    \label{tab : attack}}
\end{table}

\section{Results}
\label{sec : res}
We first show that our defended models, which add Gaussian noise to all inputs, retain low WER on this noisy but non-adversarial data. Then we report their performance against adversarial attacks. We show that they successfully recover all attacks that use near-imperceptible noise.
\subsection{Performance under Gaussian noise}
\label{sec : noisyres}

We report the performance of our model on noisy inputs (but no attack) in Table \ref{tab : clean}.
\paragraph{Augmentation and Enhancement}
We evaluate Gaussian augmentation on DeepSpeech2, and ASNR on both our architectures. Both techniques lower the word-error rate significantly under $\sigma=0.01,0.02$, with an advantage for the former. Interestingly however, combining both techniques at once (on a DeepSpeech2 model trained on noisy \textit{and} enhanced data) does not really improve results compared to using augmentation only. This suggests using ASNR enhancement as a "fallback option" in situations where retraining a model is not acceptable. This off-the-shelf method nonetheless provides competitive performance when using state-of-the-art architectures like Transformer.

\paragraph{Voting strategies} We compare all our proposed voting strategies on DeepSpeech2 outputs. As expected, majority vote brings no significant improvement over one-sentence estimation (i.e. the baseline). Logits averaging is somewhat effective; however it does not compare to ROVER, by far the best voting method even with fewer sentences.
\paragraph{Proposed models} As a consequence, we propose two smoothing-based defenses: \begin{itemize}
    \item a \textbf{trained} defense using smoothing, augmentation and ROVER. On DeepSpeech2, with $\sigma=0.01$ (resp. $0.02$) it reaches a WER of 9 (resp. 12)
    \item an \textbf{off-the-shelf} defense using smoothing, ASNR enhancement and ROVER. Applied to DeepSpeech2 this defense suffers from a higher WER of 14 (resp. 26) but is still relatively effective. With Transformer it performs much better with a WER of 8.1 (resp 15).
\end{itemize} 

\subsection{Performance under attack}
\label{sec : attackres}

In Figure \ref{fig : attack} we plot the results of our defenses and baselines against the untargeted PGD attack, as a function of the SNR used to bound the attack. They demonstrate the effectiveness of ASR smoothing: compared to the vanilla DeepSpeech2 the Word-Error-Rate improves by 20 to 50 points for all PGD attacks bounded by $SNR\geq 20dB$ for our proposed models, with both $\sigma=0.01$ and $\sigma=0.02$. When $SNR=25dB$ is sufficient to reach total denial-of-service ($WER=100$) on Deepspeech2, and $20dB$ for the MP3 compression baseline, the same feat requires SNRs of $10-15dB$ to defeat our defenses. 

Table \ref{tab : attack} reports the results of the targeted CW attack. As expected for an unbounded attack, it is partially able to break our defenses (low WER on its target), but at a high cost. The SNR it requires to achieve these results is as low as $10-14$ under $\sigma=0.01$ and $5-8$ with $\sigma=0.02$ ! This compares to $27dB$ for the undefended DeepSpeech2 and $16dB$ for the MP3 baseline. With the higher $\sigma$ in particular, the adversarial noise becomes very much audible to the human ear, even when refining it with the Imperceptible attack (Section \ref{sec : imp}). The tradeoff in clean (no attack) accuracy is fairly low for the trained defense even with $\sigma=0.02$ ($+5\%$ WER). It is higher for the untrained, off-the-shelf defense, with which using a lower deviation may be required for practical applications.

\section{Discussion}
\subsection{ROVER accuracy/time tradeoff}
\label{sec : anavote}
\begin{figure}[h]
\centering
  \centering
  \includegraphics[width=1\linewidth]{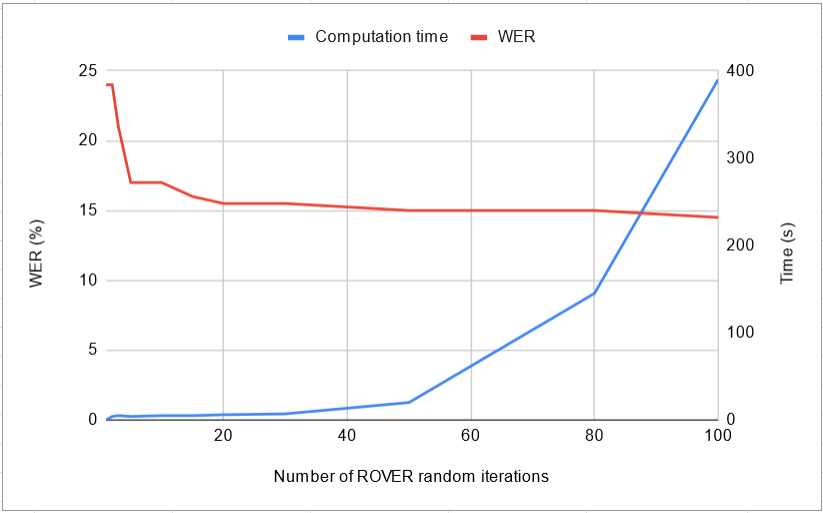}
\caption{Average ROVER Word-Error Rate and computation time (DeepSpeech2) for a single utterance when varying $N$.}
\label{fig : vote}
\end{figure}

One drawback of ROVER voting is its important time consumption when using many inputs. This may be partially due to its third-party, black-box implementation that does not use GPU computation. However, in Figure \ref{fig : vote} we show that ROVER voting time increases superlinearly with the number of samples (where averaging and counting are of course linear): this is most likely an irreducible complexity of the algorithm. Using more than 50 iterations is not practically feasible\footnote{It is, in fact, forbidden by default in the publicly available implementation}. This is why we use $N=16$ in most of our experiments: even though more iterations may bring marginal WER improvement, this value enables us to improve performance substantially while keeping the voting time negligible. Besides, 16 inputs can typically be fed to the model in one batch, thus keeping the overall computation time low. 

\subsection{Certifying randomized smoothing on ASR}
\label{sec : cert}

While our main focus in this work is to reach strong empirical performance under attack, we also show that adversarial robustness can to some extent be \textit{proven} for Speech Recognition, as is the case with classification \cite{cohen19}. We show that we can prove the following result: 

\begin{proposition} If for a sentence $s$ the randomized ASR model $f$ verifies $$\mathbb{P}[WER(f(x),s)<k]\geq p > 0.5$$ then for any noise $\norm{\delta}_2<R$: $$\mathbb{P}[WER(f(x+\delta),s)<k]> 0.5$$ where $R = \frac{\sigma}{2}(\phi^{-1}(p)-\phi^{-1}(1-p))$ and $\phi$ is  the standard Gaussian CDF.
\end{proposition}
This means that if with high probability the Gaussian-smoothed model does not deviate "too much" from a sentence $s$ in terms of WER, then the same remains true when adding a small perturbation $\delta$. This proposition can be used to write an algorithm that certifies a transcription. We defer the proof to appendix \ref{apx : proof}.

In practice such guarantees are very hard to compute: these certification algorithms demand thousands of forward passes to give results with any useful confidence margin, which on large ASR models remains an open problem.

\section{Conclusion}
\label{sec : concl}
We have proposed a state-of-the-art adversarial defense for ASR models based on randomized smoothing. It is successful against all attacks using inaudible distortion, while retaining a low error rate on natural data. To achieve strong performance under noise, we have leveraged speech enhancement methods and proposed a novel use for ASR output ensembling methods like ROVER. We successfully defend against state-of-the-art adaptive attacks and analyze the importance and limits of each component of our defense. Finally, we show that Randomized Smoothing on ASR is to some extent a provably robust defense.

This work paves the way for a thorough exploration of smoothing defenses for ASR. Practical certification, extension to other architectures and ensembling with other defenses are some areas of interest. Our approach could also be crossed with \citet{mendes} to generate noise that defends specifically against psychoacoustic-based attacks \cite{qin19}.

\section{Acknowledgements}
This material is based upon work supported by the U.S. Army Research Laboratory
and DARPA under contract HR001120C0012. Any opinions, findings and conclusions or
recommendations expressed in this material are those of the author(s)
and do not necessarily reflect the views of the the U.S. Army Research Laboratory
and DARPA.
\bibliographystyle{acl_natbib}
\bibliography{main}
\appendix
\newpage

\section{The need for adaptive attacks} \label{apx : adapt}
\subsection{Background}
In some previous works, the authors apply attacks to their proposed defense without any modification. This amounts to assuming that the attacker \textit{ignores} the existence of the defense, and tends to lead to inflated accuracy for the defender.

In the particular case of stochastic defenses, such as ours, it is a well-known fact that the gradients the attacker uses are noisy and less informative than those of undefended models, thus making gradient-based attacks less effective \cite{athalye18}. This phenomenon should not be seen as a desirable feature: rather than making adversarial examples fail to break the defense, the defense \textit{obfuscates} them but is still vulnerable to a cautious attacker, who uses an \textit{adaptive} attack rather than the vanilla attack.

One simple fix to the noisy gradient phenomenon is \textit{Expectation over Transformation} (EoT). The attacker, which has only access to the stochastic model $f(x+\epsilon)$ cannot access the gradients of the deterministic model $$\nabla_x  f(x)=\nabla_x  \mathbb{E}[f(x+\epsilon)]$$ They estimate them by applying the expectation to the gradients of the stochastic function: $$\nabla_x  f(x)\approx\mathbb{E}[\nabla_x f(x+\epsilon)]$$.

This latter quantity can be estimated with sample mean, i.e. by averaging gradients over a batch.

\subsection{Results}
All the results we report in the main paper are computed against the above adaptive attack, using gradient batches of size 16. In Table \ref{tab : adapt} we report results obtain by the PGD attack on our trained defense, with $\sigma=0.01$, with and without EoT. The WER is significantly lower using vanilla attacks, demonstrating why using adaptive attacks is necessary to correctly evaluate a defense. This also illustrates that our claims in the paper do \textbf{not} reflect obfuscation phenomena, but rather actual adversarial robustness.
\begin{table*}[h]
    \centering
    \begin{tabular}{@{\extracolsep{2pt}}ll|llllll@{}}
    \hline
       Model, Smoothing  &  Adaptive & \multicolumn{6}{c}{$\operatorname{SNR-PGD}$} \\
        &  & 35 & 30 & 25 & 20  & 15  & 10  \\
      \hline
        Trained defense, $\sigma=0.02$ & No   &  29  &  39  & 54 & 66 & 90 & 100 \\
        Trained defense, $\sigma=0.02$ & Yes  & 34 & 46 & 60 & 70 & 92 & 100 \\
       \hline
    \end{tabular}
    \caption{Word Error Rate (\%) for Our defense on the first 100 utterances of the LibriSpeech clean test set under untargeted PGD attacks using various SNR. The first line corresponds to the vanilla attack, the second uses an adaptive attack that averages gradients on 16 forward+backward passes.
    \label{tab : adapt}}
\end{table*}

\section{The ROVER voting algorithm} \label{apx : rover}
ROVER \cite{rover} works in two steps. First, it aligns all $k$ sentences word-by-word and aggregates them into one Word Transition Network (WTN), i.e. a graph where nodes represent timesteps, and edges between two timesteps are word (or silent) candidates. Alignment is done iteratively: the first sentence serves as a base WTN, then for $i=2,...,k$ ROVER merges sentence $i$ with the base WTN using Dynamic Programming tool SCLITE, using a process close to Levenstein distance: it finds the minimal cost alignment using operations of substitution, insertion and deletion. These alignment steps make use of audio alignment information as well as word and sentence scores, to output a final WTN.

At this step, ROVER votes on the aligned words using (in our version) the frequency of each word. It also accepts metrics based on word confidence, which we evaluated (using DeepSpeech2's softmax outputs as confidence scores) and found not to bring any improvement in our use case.

\section{Certifying ASR smoothing} \label{apx : proof}.
\subsection{Robustness properties for classification}
For multiclass classification, \citet{cohen19} are able to provide a robustness certificate based on the probability of the most probable class A: if  $$p_A=\mathbf{P}(f(x+\epsilon) = A)$$ and $$p_B=\max_{B \neq A}\mathbf{P}(f(x+\epsilon) = B)$$ then $g(x+\delta) = A$ for all $\norm{\delta}_2<R$ with $$R = \frac{\sigma}{2}(\phi^{-1}(p_A)-\phi^{-1}(p_B)$$ where $g$ is the smoothed classifier:
$$g(x)=argmax_{k\in \{1,2,...,m\}}\mathbf{P}(f(x+\epsilon)=k)$$ $\epsilon \sim \mathcal{N}(0,\sigma^2I)$ and $\phi$ the standard Gaussian CDF.

This result extends naturally from the binary classification case, which itself is a consequence of the Neyman-Pearson lemma \cite{neyman}. In the case of ASR, reducing the problem to binary classification is not as trivial. Below we propose such a reduction by using thresholds on the evaluation metric $d$ (typically the WER). It was brought to our attention that this idea is closely related to the more general smoothing certification on metric-space outputs proposed by \cite{Kumar21CS}. It is possible that a direct application of Center Smoothing may provide a tighter bound on the WER radius, but this remains an open question.

\subsection{Robustness properties for ASR}
Name $f$ an ASR model. We assume given an \textit{empirical target} sentence $\tilde{y}$ (different from the golden transcription $y$) and a threshold $k \in ]0,1[$. We define the binary classifier $\tilde{f}$ as: $$\tilde{f}(x) = \begin{cases}
      1 & \text{if}\ d(f(x),\tilde{y})<k \\
      0 & \text{otherwise}
    \end{cases}$$
And $\tilde{g}(x) = argmax_{c\in\{0,1\}}\mathbf{P}(\tilde{f}(x+\epsilon)=c)$. $\tilde{g}$ is informative only if $\tilde{g}(x)=1$.

By immediate application of \citet{cohen19}'s result to $\tilde{f}$ we obtain the following guarantee:
\begin{proposition}
$\tilde{g}(x+\delta)=1$ for all $\norm{\delta}_2<R$ with 
$$R = \frac{\sigma}{2}(\phi^{-1}(p)-\phi^{-1}(1-p)),\;p=\mathbf{P}(\tilde{f}(x)=1)$$
\end{proposition}

\subsubsection{Certification algorithm}
This result allows us to use on $\tilde{f}$ the CERTIFY algorithm from \citet{cohen19} (Section 3.2.2). We do not reproduce it here; the only change to our use case is that rather than generating a ``top class'' $c_A$ based on counts, we use our ROVER prediction strategy to generate the ``top transcription'' $t_A$. A policy to estimate the bound $k$ could perhaps be designed, or $k$ can simply be fixed to a value that seems reasonable with respect to applications.
\begin{proposition}
With probability at least $1 - \alpha$ over the randomness in CERTIFY, if CERTIFY returns a transcription $t_A$ and a radius $R$ (i.e.  does not abstain), then the model predicts $t_A$ at $WER\leq k$ within radius $R$ around $x$.
\end{proposition}

\subsubsection{Feasability}
Despite our good experimental results, and while we showed in this section that robustness certification is \textit{possible}, this algorithm requires estimating probabilities which is computationally infeasible using a large ASR model. As \citet{cohen19} note, certification with informative bounds and confidence requires many samples (in the tens of thousands). We are unable to run this many forward loops per sentence\footnote{Using one NVIDIA GTX 2080Ti} under reasonable time constraints. Nonetheless, these results show a possible way to reduce ASR to finite-class classification for certification purposes, and open a path towards guaranteeing the robustness of ASR models using faster architectures.

\section{Additional experiments}
In Table \ref{tab : attackfull} we report the results of both the PGD and CW attacks on our baselines and proposed models. We include models that are partially defended (without voting/augmented training/etc) to serve as an ablation study.
\begin{table*}[h]
    \centering
    \begin{tabular}{@{\extracolsep{2pt}}ll|l|llllll|lll@{}}
    \hline
       $\sigma$ & Model, Smoothing  & Nat. & \multicolumn{6}{c|}{$\operatorname{SNR-PGD}$} & \multicolumn{2}{c}{$\operatorname{CW}$}\\\cline{4-9}  \cline{10-12}
       &  &  &   35 & 30 & 25 & 20  & 15  & 10  &  GT & TGT & SNR\\
      \hline
      0.0 & Deepspeech2  &  8 & 63 & 89 & 100 & 100 & 100 & 100 & 100 & 0 & 27  \\
       & MP3 compression  & 9 & 59 & 71 & 79 & 100 & 100 & 100 & 100 & 3 & 16 \\
      \hline
        & $\operatorname{AUG}$  &  12 & 50 & 64 & 82 & 100 & 100 & 100 & 100 & 0 & 19 \\
        & $\operatorname{RS}$  &  34 & 58 & 66 & 84 & 95 & 100 & 100 & 100 & 13 & 15 \\
        & $\operatorname{RS}+\operatorname{AUG}$  &11 & 30 & 46 & 57 & 82 & 100 & 100 & 100 & 11 & 14 \\
      0.01 & $+\operatorname{RS}+\operatorname{ASNR}$  & 23 & 35 & 45 & 60 & 71 & 91 & 100 & 100 & 11 & 10  \\
        & $\operatorname{RS}+\operatorname{ROVER}$  & 29 & 51 & 64 & 81 & 93 & 100 & 100 & 100 & 7 & 15  \\
        & $\operatorname{RS}+\operatorname{\sigma-AUG}+\operatorname{ROVER}$  & 9 & 31 & 43 & 59 & 81 & 100 & 100 & 100 & 8 & 14 \\
        & $\operatorname{RS}+\operatorname{ASNR}+\operatorname{ROVER}$  & 20 & 34 & 46 & 60 & 70 & 92 & 100 & 100 & 4 & 10 \\
      \hline
        & $\operatorname{AUG}$  & 13 & 46 & 63 & 83 & 100 & 100 & 100 & 100 & 0 & 15 \\
        & $\operatorname{RS}$  & 72 & 79 & 82 & 87 & 96 & 97 & 100 & 100 & 12 & 10 \\
        & $\operatorname{RS}+\operatorname{AUG}$  &19 & 35 & 46 & 64 & 77 & 100 & 100 & 100 & 9 & 8  \\
     0.02  & $\operatorname{RS}+\operatorname{ASNR}$  &  41 & 44 & 55 & 62 & 74 & 84 & 100 & 100 & 13 & 5  \\
        & $\operatorname{RS}+\operatorname{ROVER}$  &  62 & 72 & 77 & 84 & 93 & 98 & 100 & 100 & 7 & 10 \\
        & $\operatorname{RS}+\operatorname{AUG}+\operatorname{ROVER}$  &  14 & 34 & 46 & 60 & 77 & 100 & 100 & 100 & 6 & 8 \\
        & $\operatorname{RS}+\operatorname{ASNR}+\operatorname{ROVER}$  & 36 & 40 & 46 & 56 & 71 & 83 & 100 & 100 & 6 & 5 \\
       \hline
    \end{tabular}
    \caption{Word Error Rate (\%) for Deepspeech2 on the first 100 utterances of the LibriSpeech clean test set under various attacks and defenses. $+\operatorname{AUG}$ stands for Gaussian augmentation of deviation $\sigma$ in training - the same deviation used at inference. $\operatorname{ASNR}$ means A priori SNR filtering of inputs. $+\operatorname{ROVER}$ refers to the ROVER voting strategy using 16 forward passes.
    For the PGD attack we specify the minimal SNR we use as $L_\infty$ bound. For the unbounded CW attack we report both the WER on the ground truth (GT) and the attack target (TGT), and the SNR required to achieve it. All attacks run on models using smoothing are adaptive and average gradients on 16 forward+backward passes.
    \label{tab : attackfull}}
\end{table*}

\end{document}